\documentclass[11pt]{article}

\usepackage[final]{acl}

\usepackage{times}
\usepackage{latexsym}

\usepackage[T1]{fontenc}

\usepackage[utf8]{inputenc}

\usepackage{microtype}

\usepackage{inconsolata}

\usepackage{graphicx}

\usepackage{amsmath}
\usepackage{amssymb}
\usepackage{booktabs}
\usepackage{tabularx}
\usepackage{multirow}
\usepackage{tcolorbox}
\tcbuselibrary{skins,breakable}
\usepackage{placeins}
\newcommand{\Comment}[1]{\hfill\small\textit{#1}}

\title{Navigating the Reality Gap: On-Device Continual Adaptation of ASR for Clinical Telephony}

\author{
  \textbf{Darshil Chauhan\textsuperscript{1}},
  \textbf{Adityasinh Solanki\textsuperscript{1}},
  \textbf{Vansh Patel\textsuperscript{1}},
  \textbf{Kanav Kapoor\textsuperscript{1}}, \\
  \textbf{Ritvik Jain\textsuperscript{2}},
  \textbf{Aditya Bansal\textsuperscript{2}},
  \textbf{Pratik Narang\textsuperscript{1}},
  \textbf{Dhruv Kumar\textsuperscript{1}} \\
  \textsuperscript{1}BITS Pilani, Pilani Campus, India \\
  \textsuperscript{2}Qure.ai, India \\
  \small{\texttt{\href{mailto:f20230609@pilani.bits-pilani.ac.in}{f20230609@pilani.bits-pilani.ac.in}}}
}

\begin{document}
\maketitle
\begin{abstract}
Automatic Speech Recognition (ASR) can ease clinical documentation in resource-constrained regions, but deployment is hindered by a "Reality Gap" between laboratory performance and noisy, real-world clinical telephony, compounded by strict data residency and compute constraints. We study this gap using Gram Vaani, a telephonic Hindi corpus spanning rural healthcare and agricultural helplines, as the closest publicly available proxy for clinical telephony speech, and show that a robust multilingual model (IndicWav2Vec) degrades from 11.60\% WER on clean read Hindi to 41.72\% WER on this data. We evaluate a progression of adaptation regimes, from full fine-tuning and offline Low-Rank Adaptation (LoRA) upper bounds to an on-device, stream-based continual adaptation framework in which raw audio never leaves the local device, and characterize the trade-offs between data-driven and parameter-driven stabilization strategies. Our evaluation covers both lexical accuracy (WER and CER) and semantic fidelity (BERTScore) on the target domain, alongside the retention of general-domain knowledge. Multi-domain Experience Replay (ER) yields the primary gains, improving target WER by 18.2\% relative and reducing catastrophic forgetting by 54\% compared to naive adaptation, with BERTScore reflecting consistent gains in semantic fidelity. Combining replay with Elastic Weight Consolidation based on a stabilized importance estimate (Absolute Fisher) yields the strongest retention at a small cost in plasticity. Finally, a language model spot check empirically verifies that the core mismatch lies at the acoustic level and cannot be resolved by language models alone.
\end{abstract}

\section{Introduction} \label{sec:intro}
The recent surge in Self-Supervised Learning (SSL) has propelled Automatic Speech Recognition (ASR) to near-human performance on standardized benchmarks. Foundational models like Meta's Wav2Vec 2.0 \cite{baevski2020wav2vec} and OpenAI's Whisper \cite{radford2023robust} promise a future where automated transcription can digitize patient reports, allowing clinics to improve throughput with reduced operational costs. However, for specialized, high-impact clinical domains such as rural health helplines and remote clinical services, this promise remains unfulfilled. The "reality gap", representing the disparity between clean training corpora and the chaotic, noisy, and privacy-constrained environments of real-world clinical telephony, renders these state-of-the-art models practically unusable. Consequently, ASR in these settings is best positioned as an assistive tool in clinician-in-the-loop workflows, where improvements reduce manual documentation effort rather than enable full automation.

This mismatch stems from several compounding acoustic factors. Telephonic audio in low-resource clinics is typically band-limited to 8 kHz, introducing severe channel attenuation compared to the high-fidelity speech used to pretrain foundation models. Furthermore, rural speech often features diverse regional dialects and non-standard pronunciations that are poorly represented in general-domain corpora, while ambient background noise and low-quality telephonic channels introduce non-linear distortions.

Our baseline analysis reveals that even robust multilingual models like IndicWav2Vec \cite{javed2022indicwav2vec} degrade from 11.60\% WER on clean read Hindi (Kathbath) \cite{javed2022indicsuperb} to a prohibitively high \textbf{41.72\% Word Error Rate (WER)} on rural telephony speech from Gram Vaani \cite{bhanushali2022gramvaani}, a corpus of healthcare and agricultural helpline calls that we use as the closest publicly available proxy for clinical telephony. This failure is compounded by a technical deadlock: patient data residency requirements preclude cloud-based adaptation, while rural clinics lack the compute for conventional retraining. We therefore study an on-device adaptation framework for stream-based learning in which raw patient audio never leaves the local device. This ensures data residency but is not a formal privacy guarantee.

Using Low-Rank Adaptation (LoRA) \cite{hu2022lora}, the model trains only a fraction of its parameters. We compare this setting against full fine-tuning and offline LoRA upper bounds that assume unrestricted access to the target data. However, sequential adaptation risks \textit{catastrophic forgetting} of foundational linguistic capabilities \cite{mccloskey1989catastrophic}. To counteract this, we interleave small buffers of general-domain data with the incoming stream through multi-domain experience replay \cite{chaudhry2019tiny}.

We also evaluate parameter regularization through Elastic Weight Consolidation (EWC) \cite{kirkpatrick2017overcoming}, alone and combined with replay, using an Absolute Fisher importance estimate \cite{hsu2023absolute} that is more robust than the standard squared formulation to the high-variance gradients of noisy telephony speech. We measure forgetting on two general-domain Hindi benchmarks, Kathbath and FLEURS. Since Gram Vaani does not provide named-entity or disease-name annotations, clinically targeted evaluation is not possible; we therefore quantify performance with WER and CER, complemented by BERTScore \cite{zhang2019bertscore} to capture semantic fidelity that lexical metrics miss. Our contributions are:

\begin{itemize}
    \item \textbf{On-Device Clinical Adaptation:} A continual adaptation framework for clinical telephony ASR that preserves data residency and runs on consumer-grade laptops.
    \item \textbf{Empirical Validation on Proxy Clinical Speech:} LoRA combined with multi-domain replay yields an 18.2\% relative WER reduction on rural telephony speech, averaged over three seeds and evaluated on three datasets.
    \item \textbf{Investigation of Optimization Robustness:} The interplay between data-driven replay and parameter-driven regularization, validating that linearized Fisher importance promotes more reliable convergence in noisy, multi-speaker telephony.
    \item \textbf{Quantification of the Reality Gap:} A characterization of the gap on 8kHz clinical-proxy telephony, contextualized by full fine-tuning and offline LoRA upper bounds, showing that acoustic adaptation is necessary.
\end{itemize}

\section{Related Work}

Designing an on-device, data-residency constrained, adaptive ASR pipeline for low-resource medical settings requires synthesizing advancements in self-supervised acoustic modeling, parameter-efficient adaptation, and continual learning. We address critical gaps regarding adaptation efficiency, data scarcity, and optimization stability.

\subsection{Self-Supervised Acoustic Foundations and the Domain Gap}
Low-resource speech recognition increasingly relies on Self-Supervised Learning (SSL) to leverage unlabeled audio. We build upon Wav2Vec 2.0 \citep{baevski2020wav2vec}, which achieves high data efficiency through contrastive learning; \citet{baevski2020wav2vec} showed that fine-tuning on just ten minutes of data yields competitive performance on benchmarks like Librispeech \citep{panayotov2015librispeech}. To support linguistic diversity, we utilize IndicWav2Vec \citep{javed2022indicwav2vec}, scaling the architecture to cover 40 Indian languages.

\paragraph{The Gap:} Despite these capabilities, a "usability gap" exists between laboratory benchmarks and real-world deployment. Clinical ASR errors are particularly risky because they affect medically salient information such as dosages, negations, and symptom reporting \cite{finley2018automatic, shafran2020medical}. This is exacerbated in rural healthcare by noisy telephonic audio and diverse regional dialects. Recent large-scale efforts such as NIRANTAR \cite{javed2025nirantar} have begun characterizing these challenges across 22 Indian languages, while regularizers such as EWC have been used to mitigate regional disparities in ASR \citep{trinh22_interspeech}. We utilize the Gram Vaani ASR dataset \citep{bhanushali2022gramvaani} to represent these challenges; \citet{bhanushali2022gramvaani} highlight that standard models struggle to generalize to such 8kHz telephony audio without targeted adaptation.

While signal-level enhancement strategies, such as bandwidth expansion and speech super-resolution \citep{lin2023noise, li2019audio}, offer an alternative by reconstructing high-frequency components, they introduce significant computational overhead. Deploying a separate bandwidth expansion model (often GAN or diffusion-based) alongside the ASR system contradicts the strict low-latency requirements of edge devices. Furthermore, data residency constraints preclude cloud-based services \citep{leroy2019federated}, creating a deadlock where models cannot improve because data cannot leave the local environment. While n-gram language models can improve decoding accuracy, they do not address the underlying acoustic mismatch.

For evaluation, lexical metrics such as Word Error Rate (WER) and Character Error Rate (CER) are commonly used, though they may penalize clinically acceptable synonyms or minor morphological variations. Furthermore, no universal error threshold defines clinical usability, as acceptability depends on task and acoustic conditions \citep{yoon2026optimizing}. To capture semantic fidelity, we additionally report BERTScore \citep{zhang2019bertscore}, which correlates with human judgments of semantic similarity and has been reported alongside WER and CER in recent clinical ASR systems \citep{yang2026development}.

\subsection{Parameter-Efficient Fine-Tuning (PEFT) and the Efficiency Gap}
Bridging the domain gap typically requires fine-tuning on target data. However, for Large Audio Models (LAMs), full fine-tuning is computationally prohibitive and prone to overfitting. This presents an \textit{efficiency gap}: high-end GPUs are unavailable on the resource-constrained edge devices of rural hospitals.

To address this, we adopt Low-Rank Adaptation (LoRA) \citep{hu2022lora}, which enables local fine-tuning on modest clinical hardware. This approach is essential for meeting healthcare \textbf{Data Residency} requirements, as it allows for localized, high-performance adaptation without needing the centralized data storage often required by federated or cloud-based approaches \cite{rieke2020future}. Recent work by \citet{song2024lora} on LoRA-Whisper validates this approach for multilingual ASR, demonstrating that PEFT can effectively adapt large Transformers without catastrophic interference, and LoRA has also been regularized with EWC to add new languages to ASR foundation models \citep{qian2024learn}. By constraining optimization to a low-dimensional subspace, LoRA resolves deployment challenges in data-residency constrained clinics. To understand adaptation trade-offs, we evaluate both full fine-tuning and offline LoRA fine-tuning as upper bounds.

While PEFT solves efficiency, self-improving systems face the \textit{stability gap} of Continual Learning (CL). In specialized clinical domains, annotated audio is scarce due to privacy constraints and labeling costs \citep{finley2018automatic}, and is often collected incrementally in small batches; CL in ASR has accordingly focused on adapting models sequentially without storing historical data \citep{sadhu2020continual}. Such systems face \textit{Catastrophic Forgetting}, where new training erodes previously learned capabilities \citep{mccloskey1989catastrophic}. In healthcare, ensuring model stability over time is a critical safety requirement for clinical evaluation cycles \cite{delange2021continual}. In ASR, naive updates on recent telephony transcripts lead to overfitting on specific speakers or acoustic conditions. Benchmarks like NIRANTAR \cite{javed2025nirantar} further underscore this, demonstrating that no single CL strategy currently yields consistent performance across the diverse shifts encountered in real-world Indian speech.

To mitigate this, we employ a hybrid continual learning strategy that synergizes data-level and parameter-level consolidation. First, to address the distribution shift between segments, we implement \textbf{Experience Replay (ER)} \citep{chaudhry2019tiny}. Building on recent regularized CL studies for Indic ASR \citep{adethya2025study} and episodic memory frameworks \citep{yang2022online}, we utilize a \textit{multi-domain experience replay} mechanism. While \citet{yang2022online} focused on gradient projection, we find that a direct data mixing strategy effectively mitigates catastrophic forgetting when combined with PEFT. Our buffer retains a mixture of "hard" examples (high-loss) from the target domain (Gram Vaani) and gender-balanced samples from a clean auxiliary domain (Kathbath).

Complementing this, we constrain parameter drift using \textbf{Online LoRA-EWC} \citep{kirkpatrick2017overcoming, zheng2026revisiting}, utilizing the \textbf{Absolute Fisher} importance estimation \citep{hsu2023absolute}; weight regularization has been shown to protect shared low-rank updates in continual learning \citep{zheng2026revisiting}. We estimate importance using the \textit{absolute value} of accumulated gradients ($|g|$) rather than the standard squared Fisher Information. While this linear scaling draws inspiration from the numerical stability of Memory Aware Synapses (MAS) \citep{aljundi2018memory}, Absolute Fisher explicitly ties the importance to the likelihood of the adaptation task rather than unsupervised output sensitivity. By combining this regularized signal with multi-domain replay, our approach targets catastrophic forgetting through a dual mechanism: replay maintains the data distribution, while the regularization penalty preserves critical adapter parameters.

\section{Methodology}

We study an on-device, adaptive ASR framework designed to bridge the "reality gap" within the acoustic constraints of rural clinical telephony. Our approach focuses on Continual Learning (CL), enabling a pre-trained model to adapt sequentially to incoming data streams ($\mathcal{D}_{stream}$) while maintaining strict data residency and limiting catastrophic forgetting.

\subsection{Base Acoustic Backbone}
\label{sec:method_backbone}
We utilize \textbf{IndicWav2Vec} as our acoustic backbone: a Wav2Vec 2.0 model pre-trained on diverse Indian languages, consisting of a convolutional feature encoder followed by a Transformer context network trained with the Connectionist Temporal Classification loss ($\mathcal{L}_{CTC}$). We select IndicWav2Vec as it achieves state-of-the-art performance across Indian languages unsupported by other architectures, while remaining compact enough for efficient inference and LoRA-based adaptation on consumer-grade laptops. We freeze the base model and inject trainable Low-Rank Adaptation (LoRA) matrices into the query and value projection layers, which also serves as a stability safeguard against large-scale corruption of pre-trained weights.

\subsection{Evaluation Metrics}
\label{sec:method_metrics}
We evaluate performance using both lexical accuracy and semantic fidelity. Word Error Rate (WER) and Character Error Rate (CER) measure transcription accuracy at the word and character levels, respectively, as the number of substitutions, deletions, and insertions normalized by the reference length; CER provides a finer-grained analysis of phonetic accuracy, particularly useful for agglutinative languages and dialectal variations. However, these metrics rely on exact token matching and may penalize minor orthographic variations, such as \textit{matra} errors in Devanagari, that do not significantly alter meaning. To better capture semantic fidelity, we also report BERTScore \citep{zhang2019bertscore}, which matches each token of the predicted transcription to its most similar token in the reference (and vice versa) by cosine similarity of contextual embeddings, yielding precision $P$ and recall $R$; we report their harmonic mean, $F_1 = 2PR/(P+R)$.

\subsection{Progressive Adaptation Regimes}
\label{sec:method_regimes}
We define a progression of regimes with increasing deployment constraints. Full fine-tuning on the complete target dataset serves as an approximate upper bound when no constraints are imposed; offline parameter-efficient fine-tuning (LoRA) reduces computational overhead while still requiring full data access; and stream-based continual adaptation, the primary focus of our work, considers sequential data arrival under strict data residency constraints, reflecting real-world clinical deployment.

\subsection{Data-Driven Stability: Multi-Domain Experience Replay}
\label{sec:method_er}
To mitigate catastrophic forgetting, we employ Experience Replay (ER), explicitly grounding the model's optimization trajectory with historical data. We implement a \textbf{Multi-Domain Replay} strategy that maintains a dual-source buffer $\mathcal{B}$:

\begin{itemize}
    \item \textbf{General Domain Anchor ($\mathcal{B}_{gen}$):} A gender-balanced subset of standard Hindi samples (from Kathbath) that preserves foundational phonetic robustness across diverse speaker demographics.
    \item \textbf{Target Domain History ($\mathcal{B}_{spec}$):} A sliding window of "hard" examples (high loss) and random samples from previous segments, ensuring retention of recent domain-specific adaptations.
\end{itemize}

During training, the incoming stream $\mathcal{D}_{stream}$ is concatenated with samples from $\mathcal{B}$. The optimization objective for Experience Replay is:

\begin{equation}
\begin{split}
    \mathcal{L}_{ER} = \gamma \mathbb{E}_{x \sim \mathcal{D}_{stream}} [\mathcal{L}_{CTC}(x)] \\
    + (1-\gamma) \mathbb{E}_{x \sim \mathcal{B}} [\mathcal{L}_{CTC}(x)]
\end{split}
\end{equation}

where $\gamma$ represents the mixing ratio between new data and replayed data.

\subsection{Parameter-Driven Stability: Elastic Weight Consolidation (EWC)}
\label{sec:method_ewc}
As a standalone strategy (V4.5) and a component of our hybrid approach, we implement Elastic Weight Consolidation (EWC) \citep{kirkpatrick2017overcoming} to prevent drift in critical parameters. Unlike ER, which requires data storage, EWC regularizes the model by penalizing changes to weights that are important for previous tasks. We compute the importance of each LoRA parameter $\theta_i$ using the diagonal of the Fisher Information Matrix ($F$). To ensure robust optimization against high-variance gradients, we approximate importance using accumulated \textbf{absolute gradients} rather than squared gradients (Absolute Fisher; \citealp{hsu2023absolute}), which prevents transient acoustic artifacts from drowning out meaningful weights:

\begin{equation}
    F_i = \frac{1}{N} \sum_{j=1}^{N} |\nabla_{\theta_i} \log P(y_j|x_j)|
\end{equation}

The EWC regularization loss is then applied as a quadratic penalty:

\begin{equation}
    \mathcal{L}_{EWC}(\theta) = \frac{\lambda}{2} \sum_{i} F_{i} (\theta_i - \theta^*_{i})^2
\end{equation}

where $\theta^*_i$ represents the frozen optimal parameters from the previous segment and $\lambda$ controls the regularization strength.

\subsection{Hybrid Optimization Framework (V5.1)}
\label{sec:method_hybrid}
To study the interplay between data-driven grounding and parameter-driven constraints, we evaluate a Hybrid ER + EWC framework (Strategy V5.1) that combines the multi-domain replay buffer with the EWC penalty:

\begin{equation}
    \mathcal{L}_{Total} = \mathcal{L}_{ER}(\mathcal{D}_{stream}, \mathcal{B}) + \mathcal{L}_{EWC}(\theta)
\end{equation}

where $\mathcal{L}_{ER}$ provides the gradients for acoustic adaptation and $\mathcal{L}_{EWC}$ defines a "safe" optimization region that limits catastrophic forgetting.

\section{Experiments and Results}
\label{sec:experiments}

We evaluate adaptation to the target domain (Gram Vaani) and retention of general linguistic knowledge (Kathbath and FLEURS). All results are reported as mean $\pm$ standard deviation over three random seeds.

\subsection{Datasets}
\begin{itemize}
    \item \textbf{Gram Vaani (Clinical Helpline Proxy)} \citep{bhanushali2022gramvaani}: This dataset consists of rural telephonic speech from healthcare and agricultural helplines (originally 8kHz, upsampled to 16kHz as required by IndicWav2Vec; all reported results, including the baseline, are computed on the upsampled audio). Capturing rural, noisy, conversational telephony, it serves as the closest available proxy for clinical telephony audio. We evaluate on the official 3-hour Evaluation set (GV\_Eval\_3h). To simulate a continual learning scenario, the 103 hours of training data are partitioned into sequential segments processed one by one, mimicking a live data stream.
    \item \textbf{Kathbath (Source Domain)}: A high-quality, read speech dataset representing standard Hindi \citep{javed2022indicsuperb}. A subset of the training set (approx. 25,800 samples) forms the pool from which general-domain replay samples are drawn, and the complete validation set (3,151 samples) is used exclusively to measure catastrophic forgetting.
    \item \textbf{FLEURS (General Domain)} \citep{conneau2023fleurs}: We use the labeled Hindi (\texttt{hi\_in}) test split as a second benchmark to evaluate generalization and measure catastrophic forgetting under domain shift.
\end{itemize}

\subsection{Continual Adaptation Paradigms}
We investigate four continual learning paradigms, contextualized by upper-bound regimes. All experiments start from \texttt{IndicWav2Vec}, with a baseline of 41.72\% WER on Gram Vaani, 11.60\% on Kathbath, and 18.16\% on FLEURS. Every model is trained for 3 epochs: per segment for the streaming strategies, and on the full dataset for the upper bounds. LoRA hyperparameters were selected based on prior literature and preliminary validation experiments, and were kept consistent across the replay, EWC, and hybrid strategies to enable fair comparison (Appendix~\ref{sec:appendix_setup}).

\subsubsection{Upper-Bound Reference: Full and Offline LoRA Fine-Tuning}
Following Section~\ref{sec:method_regimes}, full fine-tuning (FT) updates the entire model on the complete target training set, while offline LoRA trains adapters on the complete dataset with (i) rank 16, $\alpha=32$ and (ii) rank 24, $\alpha=48$. \textbf{Outcome:} Full fine-tuning reaches \textbf{30.90\%} target WER but nearly doubles general-domain error (22.88\% on Kathbath). Offline LoRA reaches 34.90\% ($r=16$) and 34.62\% ($r=24$) target WER, with substantial forgetting (20.25\% and 21.11\% on Kathbath).

\subsubsection{Paradigm 1: Naive Continual Fine-tuning (Baseline)}
The model is fine-tuned sequentially on the incoming stream without explicit regularization, using LoRA rank 16 ($\alpha=32$) and 19 segments of 2,000 samples. \textbf{Outcome (V1.1):} Target WER drops to \textbf{35.02\%}, but general-domain error increases by an absolute \textbf{5.81\%} on Kathbath (5.01\% on FLEURS), indicating severe catastrophic forgetting.

\subsubsection{Paradigm 2: Experience Replay (ER) Strategies}
We investigate two ER configurations, increasing LoRA capacity to rank 24 ($\alpha=48$) to accommodate the dual-task nature of rehearsal.
\begin{itemize}
    \item \textbf{Single-Domain ER (V2.1)}: Replays 400 samples from the target-domain history per segment using a "60\% hard, 40\% random" selection strategy, where hard samples have a loss above $1.2\times$ the segment average. Each segment combines 1,600 new samples with 400 replay samples (24 segments of 2,000 samples).
    \item \textbf{Multi-Domain ER (V3.1)}: Replays a balanced buffer of 300 target samples (same 60:40 split) and 300 general-domain (Kathbath) samples, so each segment contains 1,600 new and 600 replay samples. This strategy achieved our lowest continual target WER of \textbf{34.11\%} and reduced forgetting by \textbf{54\%} on Kathbath (56\% on FLEURS) relative to the naive baseline.
\end{itemize}

\subsubsection{Paradigm 3: Elastic Weight Consolidation (EWC)}
We examine whether parameter regularization alone, using the Absolute Fisher penalty (Section~\ref{sec:method_ewc}) with $\lambda=10$ (larger values were overly conservative or prevented adaptation; Section~\ref{sec:ablation_lambda}), can protect foundational knowledge without replaying data. V4.5 adds this penalty to the naive setup, using the same 19 segments of 2,000 samples as V1.1. \textbf{Outcome (V4.5):} V4.5 achieves a competitive \textbf{34.55\% WER} on the target domain, but mitigates forgetting less efficiently than the ER strategies (+4.76\% on Kathbath) and shows the highest variance across seeds.

\subsubsection{Paradigm 4: Hybrid ER + EWC Optimization}
V5.1 combines the V3.1 replay buffer with the EWC penalty ($\lambda=10$; Section~\ref{sec:method_hybrid}). \textbf{Outcome (V5.1):} This yields our most stable model, with the lowest forgetting: \textbf{14.15\%} WER on Kathbath (+2.55\%) and \textbf{20.27\%} on FLEURS (+2.11\%). This stability comes at a slight cost to plasticity, with a final target WER of \textbf{34.66\%}.

\section{Analysis and Discussion}
\label{sec:analysis}

\subsection{Comparative Performance Summary}
Table~\ref{tab:main_results} summarizes the final metrics. All continual paradigms bridge the "Reality Gap", reducing target WER by 16.1--18.2\% relative to the baseline and converging to 34--35\% WER, and forgetting trends on FLEURS closely mirror those on Kathbath. Streaming strategies reach target WER comparable to the offline LoRA upper bounds (34.11--35.02\% vs.\ 34.62--34.90\%) while retaining substantially more general-domain knowledge. BERTScore closely follows WER across datasets, indicating consistent gains in semantic fidelity, although it is not sensitive to clinically critical distinctions such as negations or numerical magnitude errors.

\begin{table*}[t]
\centering
\small
\begin{tabular}{lll|ccc}
\toprule
\textbf{Paradigm} & \textbf{Strategy} & \textbf{Dataset} & \textbf{WER (\%)} & \textbf{CER (\%)} & \textbf{BERTScore (\%)} \\
\midrule
\multirow{9}{*}{\shortstack{Upper\\Bound}} & \multirow{3}{*}{Full Finetune}
& Gram Vaani & \underline{\textbf{30.90 $\pm$ 0.10}} & \underline{\textbf{13.20 $\pm$ 0.09}} & \underline{\textbf{89.60 $\pm$ 0.06}} \\
& & Kathbath & 22.88 $\pm$ 0.03 & 6.73 $\pm$ 0.01 & 93.86 $\pm$ 0.01 \\
& & FLEURS & 28.21 $\pm$ 0.04 & 10.17 $\pm$ 0.03 & 91.57 $\pm$ 0.04 \\
\cmidrule{2-6}
& \multirow{3}{*}{LoRA ($r$=16, $\alpha$=32)}
& Gram Vaani & 34.90 $\pm$ 0.05 & 14.66 $\pm$ 0.05 & 88.26 $\pm$ 0.02 \\
& & Kathbath & 20.25 $\pm$ 0.05 & 5.61 $\pm$ 0.03 & 94.51 $\pm$ 0.03 \\
& & FLEURS & 26.35 $\pm$ 0.12 & 9.13 $\pm$ 0.02 & 92.08 $\pm$ 0.03 \\
\cmidrule{2-6}
& \multirow{3}{*}{LoRA ($r$=24, $\alpha$=48)}
& Gram Vaani & 34.62 $\pm$ 0.09 & 14.58 $\pm$ 0.09 & 88.38 $\pm$ 0.04 \\
& & Kathbath & 21.11 $\pm$ 0.05 & 5.99 $\pm$ 0.02 & 94.20 $\pm$ 0.03 \\
& & FLEURS & 26.94 $\pm$ 0.12 & 9.53 $\pm$ 0.04 & 91.85 $\pm$ 0.04 \\
\midrule
\multirow{3}{*}{Baseline} & \multirow{3}{*}{IndicWav2Vec}
& Gram Vaani & 41.72 $\pm$ 0.00 & 19.05 $\pm$ 0.00 & 86.26 $\pm$ 0.00 \\
& & Kathbath & \underline{\textbf{11.60 $\pm$ 0.00}} & \underline{\textbf{3.41 $\pm$ 0.00}} & \underline{\textbf{97.28 $\pm$ 0.00}} \\
& & FLEURS & \underline{\textbf{18.16 $\pm$ 0.00}} & \underline{\textbf{6.38 $\pm$ 0.00}} & \underline{\textbf{94.74 $\pm$ 0.00}} \\
\midrule
\multirow{3}{*}{Naive} & \multirow{3}{*}{V1.1}
& Gram Vaani & 35.02 $\pm$ 0.08 & 14.38 $\pm$ 0.22 & 88.37 $\pm$ 0.08 \\
& & Kathbath & 17.41 $\pm$ 0.17 & 4.65 $\pm$ 0.02 & 95.47 $\pm$ 0.03 \\
& & FLEURS & 23.17 $\pm$ 0.17 & 7.99 $\pm$ 0.04 & 93.22 $\pm$ 0.05 \\
\midrule
\multirow{6}{*}{ER} & \multirow{3}{*}{V2.1 (Single Domain)}
& Gram Vaani & 34.24 $\pm$ 0.05 & 14.28 $\pm$ 0.11 & 88.77 $\pm$ 0.04 \\
& & Kathbath & 15.06 $\pm$ 0.09 & 4.05 $\pm$ 0.03 & 96.61 $\pm$ 0.01 \\
& & FLEURS & 21.09 $\pm$ 0.09 & 7.32 $\pm$ 0.06 & 94.45 $\pm$ 0.04 \\
\cmidrule{2-6}
& \multirow{3}{*}{V3.1 (Multi Domain)}
& Gram Vaani & \textbf{34.11 $\pm$ 0.02} & 14.22 $\pm$ 0.05 & \textbf{88.85 $\pm$ 0.04} \\
& & Kathbath & 14.27 $\pm$ 0.02 & 3.87 $\pm$ 0.01 & 96.78 $\pm$ 0.01 \\
& & FLEURS & 20.36 $\pm$ 0.09 & 7.10 $\pm$ 0.02 & \textbf{94.65 $\pm$ 0.04} \\
\midrule
\multirow{3}{*}{EWC} & \multirow{3}{*}{V4.5}
& Gram Vaani & 34.55 $\pm$ 0.38 & \textbf{14.21 $\pm$ 0.10} & 88.65 $\pm$ 0.22 \\
& & Kathbath & 16.36 $\pm$ 1.04 & 4.37 $\pm$ 0.31 & 95.97 $\pm$ 0.55 \\
& & FLEURS & 22.20 $\pm$ 1.02 & 7.70 $\pm$ 0.36 & 93.75 $\pm$ 0.61 \\
\midrule
\multirow{3}{*}{Hybrid} & \multirow{3}{*}{V5.1 (ER + EWC)}
& Gram Vaani & 34.66 $\pm$ 0.07 & 14.46 $\pm$ 0.02 & 88.69 $\pm$ 0.04 \\
& & Kathbath & \textbf{14.15 $\pm$ 0.01} & \textbf{3.85 $\pm$ 0.01} & \textbf{96.80 $\pm$ 0.01} \\
& & FLEURS & \textbf{20.27 $\pm$ 0.04} & \textbf{7.04 $\pm$ 0.01} & 94.62 $\pm$ 0.01 \\
\bottomrule
\end{tabular}
\caption{Performance of the investigated adaptation regimes on the target (Gram Vaani), source (Kathbath), and general (FLEURS) domains, without LM decoding. Results are mean $\pm$ std over 3 seeds. Bold marks the best continual strategy per dataset and metric; bold underlined values are reference results: the best target-domain performance (full fine-tuning) and the unadapted general-domain performance (baseline). Forgetting corresponds to the increase in Kathbath/FLEURS WER over the IndicWav2Vec baseline.}
\label{tab:main_results}
\end{table*}

\begin{figure*}[t!]
  \centering
  \begin{minipage}{\textwidth}
    \centering
    \includegraphics[width=\textwidth]{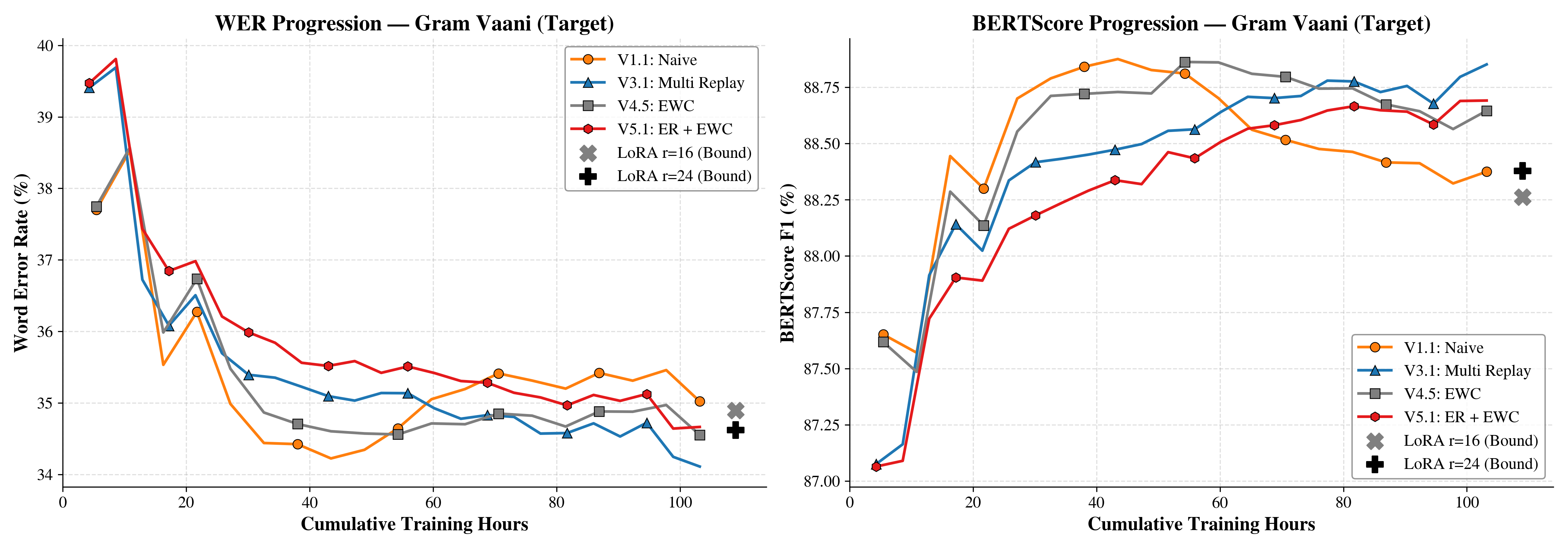}
    \caption{Progression of WER and BERTScore on the target domain (Gram Vaani) over cumulative training hours for the Naive (V1.1), Multi-Domain ER (V3.1), EWC (V4.5), and Hybrid (V5.1) strategies, with offline LoRA upper bounds for reference.}
    \label{fig:adaptation_progression}
  \end{minipage}
  \vspace{0.3cm}
  \begin{minipage}{\textwidth}
    \centering
    \includegraphics[width=\textwidth]{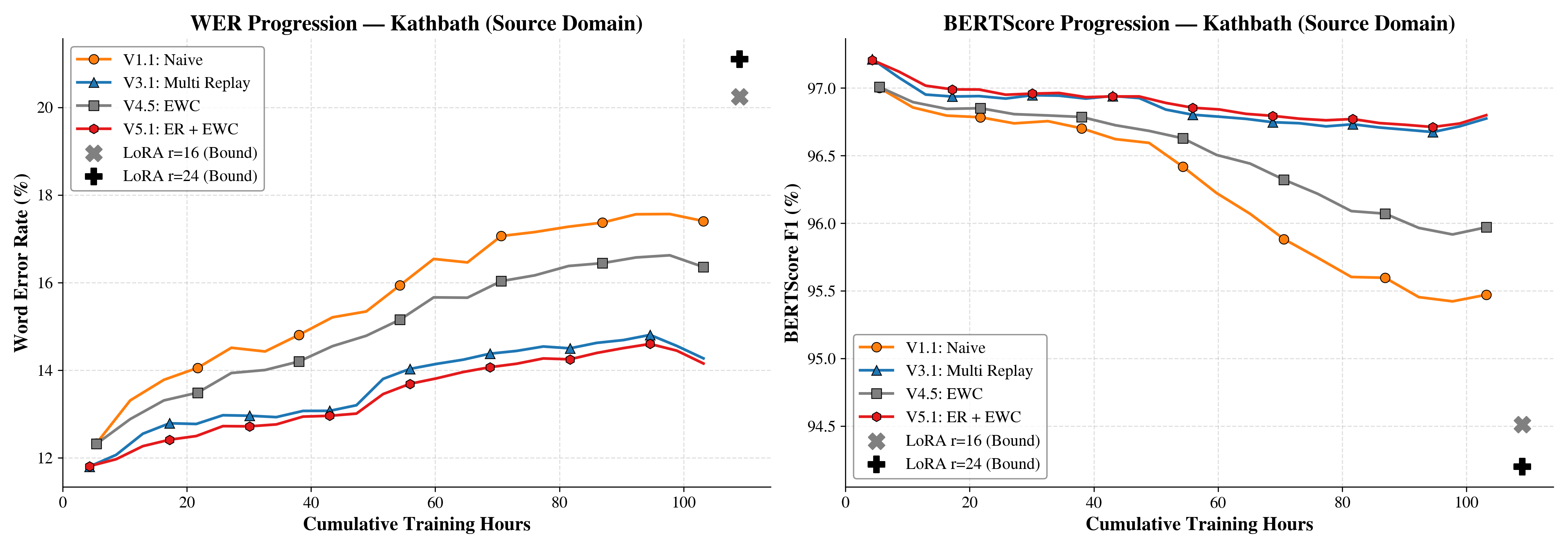}
    \caption{Catastrophic forgetting on the source domain (Kathbath): progression of WER and BERTScore as adaptation proceeds over Gram Vaani, with offline LoRA upper bounds for reference.}
    \label{fig:forgetting_curves_kathbath}
  \end{minipage}
  \vspace{0.3cm}
  \begin{minipage}{\textwidth}
    \centering
    \includegraphics[width=\textwidth]{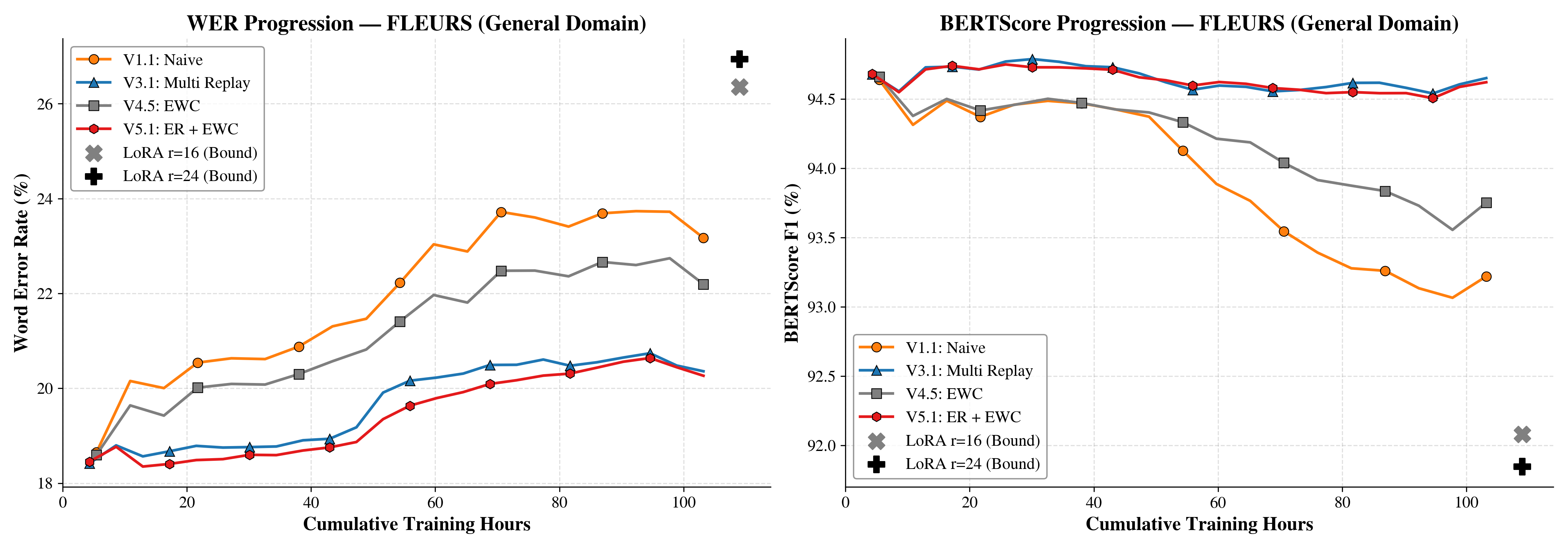}
    \caption{Catastrophic forgetting on the general domain (FLEURS): progression of WER and BERTScore as adaptation proceeds over Gram Vaani, with offline LoRA upper bounds for reference.}
    \label{fig:forgetting_curves_fleurs}
  \end{minipage}
\end{figure*}

\begin{figure*}[t]
\centering
\begin{minipage}{0.48\textwidth}
    \centering
    \includegraphics[width=\linewidth]{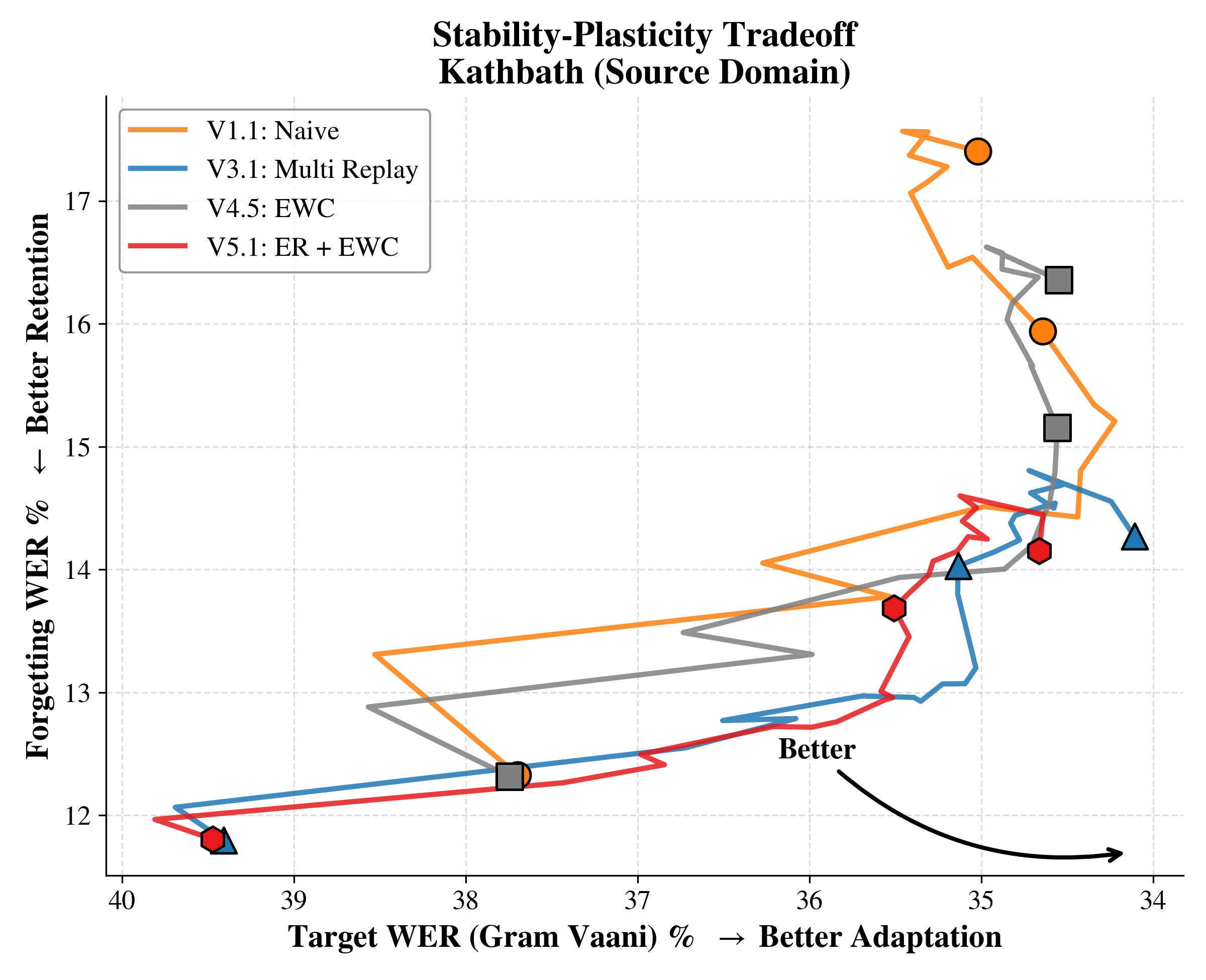}
    \caption{Stability-plasticity trajectories on Kathbath (source domain). Curves trace per-segment means over three seeds; markers denote the first, middle, and final segments.}
    \label{fig:pareto_kathbath}
\end{minipage}
\hfill
\begin{minipage}{0.48\textwidth}
    \centering
    \includegraphics[width=\linewidth]{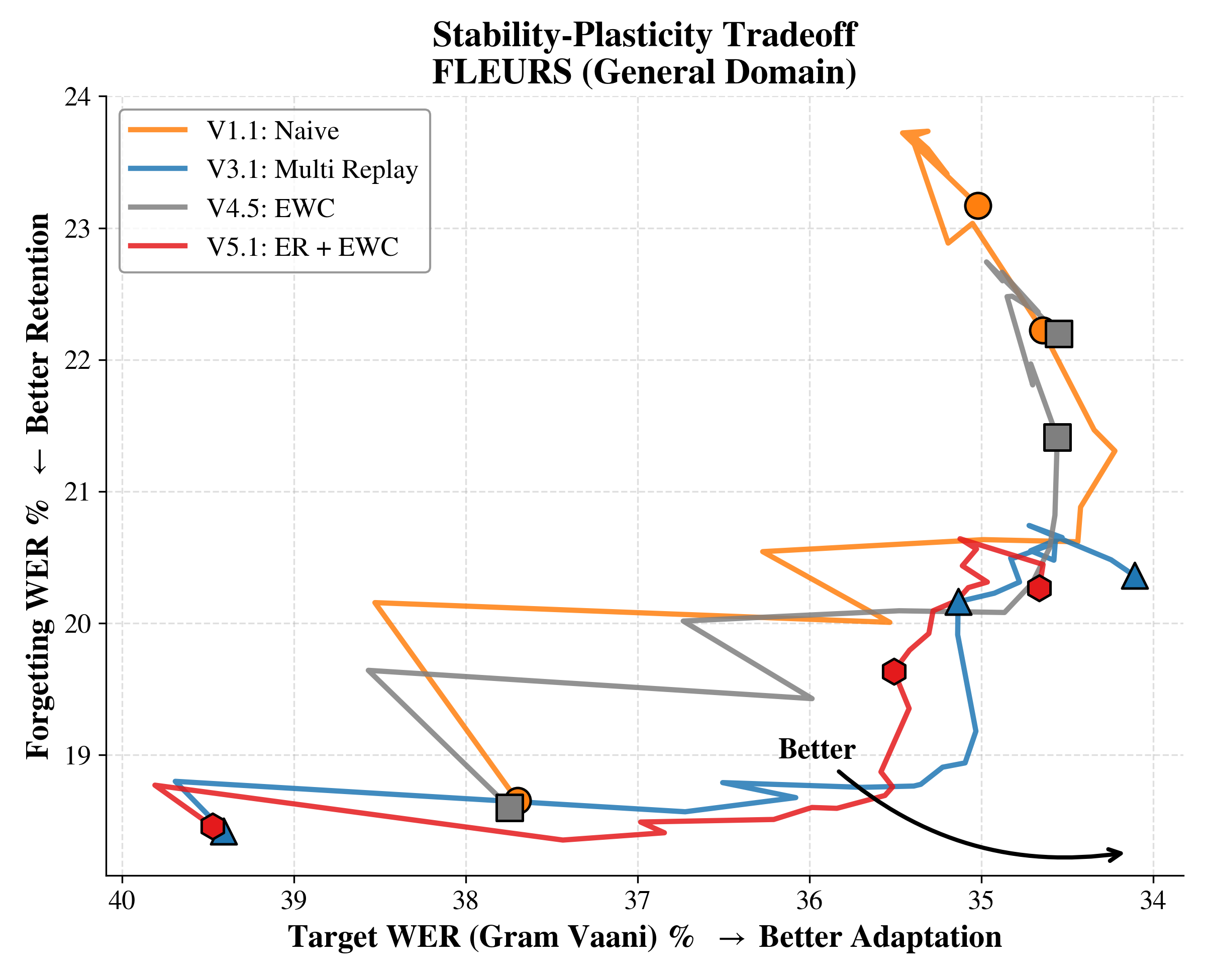}
    \caption{Stability-plasticity trajectories on FLEURS (general domain). Curves trace per-segment means over three seeds; markers denote the first, middle, and final segments.}
    \label{fig:pareto_fleurs}
\end{minipage}
\end{figure*}

\subsection{Stability-Plasticity Pareto Efficiency}
Figures~\ref{fig:pareto_kathbath} and~\ref{fig:pareto_fleurs} plot retention against adaptation, where movement toward the bottom-right corresponds to improved adaptation with minimal forgetting. The replay-based trajectories show a clear elbow near 35\% target WER, beyond which further reductions incur disproportionately high stability costs. Multi-Domain ER (V3.1) navigates this trade-off most efficiently, reaching the lowest target WER at moderate forgetting, while the Naive (V1.1) and EWC-only (V4.5) trajectories reach comparable target WER at considerably higher forgetting. This behavioral ceiling suggests that for a given model architecture and domain mismatch, there exists an optimal frontier of acoustic adaptation. Crucially, the V3.1 trajectory has not plateaued, suggesting that further gains are possible but require significantly more data or linguistic context to unlock.

\subsection{Mechanism Analysis: Data vs. Parameter Anchors}
Our investigation reveals a fundamental difference between data-driven and parameter-driven regularization in lightweight models.
\begin{itemize}
    \item \textbf{Data Anchors (ER)}: Replaying Kathbath samples provides an explicit gradient signal toward a generalized linguistic center, making replay the most efficient stabilizer.
    \item \textbf{Parameter Anchors (EWC)}: While EWC protects weights, it is "acoustically blind", penalizing drift without reconciling new acoustic features. This is evidenced by V2.1 (ER) providing superior stabilization (+3.46\% forgetting on Kathbath, +2.93\% on FLEURS) compared to V4.5 (EWC; +4.76\% and +4.04\%) despite similar target performance (34.24\% vs.\ 34.55\%).
    \item \textbf{The LoRA-Regularization Bottleneck}: We observe a conflict in the Hybrid model (V5.1), where combining LoRA's low-rank space with EWC's parameter-level penalties overly restricts optimization paths, leading to higher target WER (34.66\% vs.\ 34.11\% for V3.1).
\end{itemize}

\subsection{Convergence Dynamics}
While the Naive baseline (V1.1) exhibits erratic trajectories, the ER and Hybrid paradigms demonstrate controlled, largely monotonic trends (Figure~\ref{fig:adaptation_progression}; per-strategy progressions are in Appendix~\ref{sec:appendix_results}). These trends are consistent across seeds: replay-based strategies show the lowest seed variance (e.g., $\pm$0.02 target WER for V3.1), whereas EWC-only adaptation (V4.5) is the most seed-sensitive ($\pm$0.38 target WER, $\pm$1.04 Kathbath WER). This stability is a critical requirement for clinical deployment, ensuring the system does not regress after extended use. However, we observe a "soft floor" around 34\% WER, which we characterize as the \textbf{Acoustic Bottleneck of Clinical Telephony}. The mismatch between 8kHz telephony and the 16kHz model expectation creates a ceiling for acoustic-only adaptation. 

\subsection{The Necessity of Acoustic Adaptation}
To verify that acoustic adaptation is a \textit{prerequisite} for clinical deployment, we paired the baseline and adapted models with a domain-specific 4-gram LM (Table~\ref{tab:lm_results}). The LM improves both, but its gains remain limited without acoustic adaptation: the baseline+LM reaches only 35.21\% WER, compared to 30.32\% for V3.1+LM. With the LM, full fine-tuning reaches 29.03\% WER (at the cost of substantial forgetting), comparable to prior work on Gram Vaani (30.3\% on the Development set with Conformer/TDNN models and external LMs; \citealp{patel2022gramvaani}), and the continual strategies approach this performance under streaming constraints. We keep the LM out of the adaptation loop so that it does not mask the acoustic adaptation dynamics we study; our results without an LM are therefore a conservative estimate of achievable WER.

\begin{table}[t]
\centering
\resizebox{\columnwidth}{!}{%
\begin{tabular}{lcc}
\toprule
\textbf{Model (with LM)} & \textbf{WER (\%)} & \textbf{BERTScore (\%)} \\
\midrule
IndicWav2Vec (Baseline) & 35.21 $\pm$ 0.00 & 87.94 $\pm$ 0.00 \\
Full Finetune & \textbf{29.03 $\pm$ 0.12} & \textbf{90.14 $\pm$ 0.04} \\
LoRA Finetune ($r$=16) & 31.06 $\pm$ 0.07 & 89.45 $\pm$ 0.03 \\
V3.1 (Multi-Domain ER) & 30.32 $\pm$ 0.03 & 89.91 $\pm$ 0.02 \\
V5.1 (ER + EWC) & 30.46 $\pm$ 0.03 & 89.83 $\pm$ 0.05 \\
\bottomrule
\end{tabular}%
}
\caption{LM spot check on Gram Vaani: effect of a domain-specific 4-gram LM on different adaptation regimes. Results are mean $\pm$ std over 3 seeds. Acoustic adaptation remains essential even with LM assistance.}
\label{tab:lm_results}
\end{table}

\subsection{Ablation Studies}

\subsubsection{Sensitivity to EWC Regularization (\texorpdfstring{$\lambda$}{lambda})}
\label{sec:ablation_lambda}
With high regularization ($\lambda \geq 10^3$), the EWC constraint dominated the gradients, effectively "freezing" the LoRA parameters and preventing meaningful adaptation to the noisy telephony domain. An intermediate value of $\lambda=10^2$ allowed adaptation but remained overly conservative. A lower value of $\lambda=10$ (V4.5) provided the best balance between adaptation and parameter protection, and linear (absolute) Fisher estimation yielded more stable optimization than higher-order approximations.

\subsubsection{LR Warmup and Convergence Speed}
Compared to a conservative 100-step learning rate warmup, an aggressive 10-step schedule significantly accelerated convergence early in each segment without increasing catastrophic forgetting ($<0.05\%$ difference across all runs), suggesting that LoRA-based adaptation can safely use high initial learning rates to escape local minima from previous segments.

\FloatBarrier
\section{Conclusion}
In this work, we investigated the "Reality Gap" that hinders the deployment of state-of-the-art ASR in rural clinical settings, quantifying it at 41.72\% WER on a telephony proxy for clinical speech. Framed as an empirical study of adaptation dynamics under data residency constraints rather than a clinical validation, our comparison of four continual learning paradigms against full fine-tuning and offline LoRA upper bounds shows that Multi-Domain Experience Replay provides the best balance of plasticity and stability. While full fine-tuning achieves the strongest target performance, replay-based streaming adaptation reaches target WER comparable to offline LoRA while forgetting substantially less, improving target WER by 18.2\% relative and reducing forgetting by 54\% over naive adaptation, consistently across three seeds and two general-domain benchmarks. These findings provide a foundation for reliable, self-improving clinical ASR and its extension across Indic languages.

\section*{Privacy, Ethics, and Clinical Safety Considerations}
\label{sec:ethics}
The deployment of ASR in frontline healthcare settings introduces unique ethical and safety considerations which our framework explicitly addresses.

\paragraph{Data Governance and Residency:} Our primary focus on on-device adaptation is a direct response to healthcare data residency requirements. By ensuring that raw patient audio never leaves the local environment, we mitigate the risk of centralized data breaches and support compliance with data protection requirements in various jurisdictions \citep{leroy2019federated}. We emphasize that data residency alone is not a formal privacy guarantee: we do not study differential privacy, secure aggregation, or defenses against model-based attacks such as membership inference or model inversion.

\paragraph{Clinical Safety and Error Risks:} While our framework significantly reduces WER, automated transcription in medical contexts is never risk-free. ASR errors in medically salient keywords (e.g., negations or dosage numbers) can lead to serious diagnostic errors. We advocate for a \textbf{Clinician-in-the-Loop} model, where our adaptive system serves as a draft-generation tool that reduces manual workload while maintaining human oversight for final clinical validation. Improved accuracy reduces, but does not eliminate, clinical risk; moreover, our experiments use Gram Vaani as the closest available proxy for clinical telephony, and should not be read as clinical validation.

\paragraph{Algorithmic Bias and Equity:} By prioritizing adaptation on rural dialects and low-bandwidth telephony, this work contributes to healthcare equity. Standard ASR models often fail on marginalized populations due to accent mismatch; our framework allows localized clinics to "correct" these biases by adapting specifically to their local community's speech patterns.

\paragraph{Controlled Adaptation and Model Drift:} Continual adaptation introduces the risk of unintended model drift. Our use of experience replay and EWC-based regularization helps limit forgetting of general language knowledge while adapting to local conditions, which is particularly important in clinical settings where robustness and consistency are critical.

\section*{Limitations}
While our framework demonstrates significant potential for clinical deployment, several limitations remain:
\begin{enumerate}
    \item \textbf{Metrics as a Clinical Proxy:} Our evaluation relies on WER and CER, complemented by BERTScore. While BERTScore captures semantic similarity beyond exact token matching, none of these metrics is sensitive to clinically critical distinctions such as negations, numerical magnitude errors, or medical entity substitutions, and none explicitly weights the clinical significance of errors: an error in a patient's name may be less critical than an error in a medication dosage. Since Gram Vaani does not provide named-entity or disease-name annotations, clinically targeted evaluation (e.g., accuracy on disease names) was not possible. Developing medical-concept-aware metrics, and validating utility through deployment and clinician feedback, remain essential future directions.
    \item \textbf{Proxy Data:} We use Gram Vaani, a corpus of rural healthcare and agricultural helpline calls, as the closest available proxy for clinical telephony. Although it reflects the channel and dialectal conditions of rural telephony, the content of real clinical conversations may differ. Collecting and evaluating on real clinical telephony data requires ethics approval and patient consent frameworks beyond the present scope; it remains a primary direction for future work.
    \item \textbf{Reliance on Clinician Supervision:} Our continual learning setup assumes a stream of corrected transcripts (e.g., from medical professionals). In under-staffed rural clinics, this supervision may be sparse or delayed, necessitating future work into uncertainty-aware or semi-supervised adaptation.
    \item \textbf{Acoustic-Only Training:} While we validated the necessity of acoustic adaptation, we did not integrate Language Models (LMs) into the training loop (e.g., via shallow fusion during replay), as doing so would conflate acoustic and linguistic adaptation signals. Capturing the synergy between acoustic and linguistic adaptation is critical for handling specialized medical terminology.
    \item \textbf{Language, Dialect, and Backbone Scope:} Our experiments focus on rural Hindi dialects using a single acoustic backbone, IndicWav2Vec, chosen for its coverage of Indian languages. Evaluating additional ASR architectures would further establish robustness and remains important future work. Our immediate aim, however, is to extend the study to other Indic languages, including tonal and Dravidian families, as annotated telephonic corpora become available, and for these languages IndicWav2Vec remains the most suitable backbone.
    \item \textbf{Deployment Hardware:} In this work, "edge device" refers to consumer-grade laptops with entry-level GPUs, not mobile phones or embedded hardware. Reported runtimes (Appendix~\ref{sec:appendix_hardware}) are approximate and vary across environments.
\end{enumerate}

\section*{Acknowledgments}
The authors wish to acknowledge the use of large language models in improving the presentation and grammar of this paper. The paper remains an accurate representation of the authors' underlying contributions.

\bibliography{custom}

\clearpage
\appendix

\section{Experimental Setup and Reproducibility}
\label{sec:appendix_setup}

To ensure the reproducibility of our findings, we provide the full configuration details for all experimental paradigms. All models were trained for \textbf{3 epochs per data segment} to ensure local convergence, and all results are averaged over three random seeds. All results, including the baseline, were obtained under a single, consistent set of software package versions.

\subsection{Hyperparameter Configurations}
Table~\ref{tab:hyperparams} details the parameters used, utilizing a split-table format to distinguish between universal settings and those tailored to specific strategies. LoRA hyperparameters were selected based on prior literature and preliminary validation experiments; the replay, EWC, and hybrid strategies share an identical LoRA configuration ($r=24$, $\alpha=48$) to enable fair comparison.

\begin{table}[t]
\centering
\small
\begin{tabularx}{\columnwidth}{@{}l>{\raggedright\arraybackslash}X@{}}
\toprule
\textbf{Parameter} & \textbf{Value} \\
\midrule
Optimizer & AdamW \\
Weight Decay & $0.01$ \\
Warmup Steps & 10 (Aggressive) \\
LoRA Target Modules & query, value (Attention blocks) \\
Batch Size (Effective) & 64 \\
Max Audio Duration & 30.0 seconds \\
Training Epochs & 3 per segment \\
\midrule
\multicolumn{2}{@{}l}{\textbf{Upper Bound Strategies}} \\
Full Finetune & Base LR $= 1 \times 10^{-4}$, full model fine-tuned on combined target corpus \\
LoRA ($r$=16, $\alpha$=32) & Base LR $= 3 \times 10^{-4}$, LoRA adapters only \\
LoRA ($r$=24, $\alpha$=48) & Base LR $= 3 \times 10^{-4}$, LoRA adapters only \\
\midrule
\multicolumn{2}{@{}l}{\textbf{Continual Learning Strategies}} \\
V1.1 Naive & Base LR $= 3 \times 10^{-4}$, $r=16$, $\alpha=32$ \\
V2.1 Single-Domain ER & Base LR $= 3 \times 10^{-4}$, $r=24$, $\alpha=48$, Replay buffer: 400 target samples \\
V3.1 Multi-Domain ER & Base LR $= 3 \times 10^{-4}$, $r=24$, $\alpha=48$, Replay buffer: 300 target + 300 general \\
V4.5 EWC & Base LR $= 3 \times 10^{-4}$, $r=24$, $\alpha=48$, $\lambda=10$ \\
V5.1 Hybrid & Base LR $= 3 \times 10^{-4}$, $r=24$, $\alpha=48$, $\lambda=10$, Replay buffer: 300 target + 300 general \\
\bottomrule
\end{tabularx}
\caption{Comprehensive hyperparameter settings. $r$ and $\alpha$ represent the LoRA rank and scaling factor, respectively. $\lambda$ denotes the EWC regularization strength.}
\label{tab:hyperparams}
\end{table}

\subsection{Algorithm and Buffer Management}
The technical contribution of our framework lies in the efficient integration of LoRA, linearized EWC, and prioritized experience replay. Algorithm 1 below details the localized adaptation loop, and Algorithm 2 describes our multi-domain buffer management strategy.

\begin{figure}[t]
\begin{tcolorbox}[
    colback=white,
    colframe=black,
    boxrule=0.8pt,
    arc=0pt,
    left=4pt, right=4pt, top=4pt, bottom=4pt,
    label=alg:adaptation
]
\small
\textbf{Algorithm 1: LoRA-based Hybrid Adaptation Loop}\\
\vspace{0.1cm}
\textbf{Require:} Base model $\theta_{base}$, Replay buffer $\mathcal{B}$, Regularization $\lambda$\\
\textbf{1:} Inject LoRA adapters $\theta_{0}$ into frozen $\theta_{base}$; set importance $F \gets \mathbf{0}$\\
\textbf{2:} \textbf{for} segment $k = 1, \dots, K$ \textbf{do}\\
\textbf{3:} \quad Receive clinical stream $\mathcal{D}_{k}$\\
\textbf{4:} \quad $\mathcal{D}_{train} \leftarrow \text{Sample}(\mathcal{D}_{k}) \cup \text{Sample}(\mathcal{B})$\\
\textbf{5:} \quad \textbf{Adaptation Step:}\\
\quad \quad $\theta_{k} \leftarrow \text{argmin}_{\theta} \mathcal{L}_{CTC}(\mathcal{D}_{train})$ \\
\quad \quad \quad $+ \frac{\lambda}{2} \sum F_{i} (\theta_k - \theta_{k-1}^*)^2$\\
\textbf{6:} \quad \textbf{Importance Estimation:}\\
\quad \quad $F_{new} \gets \frac{1}{|\mathcal{D}_{train}|} \sum_{x \in \mathcal{D}_{train}} |\nabla_{\theta} \mathcal{L}_{CTC}(x)|$\\
\textbf{7:} \quad \textbf{Asynchronous Consolidation:}\\
\quad \quad $F \gets \frac{F \times (k-1) + F_{new}}{k}$\\
\textbf{8:} \quad $\theta_{k}^* \leftarrow \text{detach}(\theta_{k})$ \Comment{Checkpointing}\\
\textbf{9:} \quad $\mathcal{B} \leftarrow \text{UpdateBuffer}(\mathcal{D}_{k}, \mathcal{L}_{inst})$\\
\textbf{10:} \textbf{end for}
\end{tcolorbox}
\end{figure}

\begin{figure}[t]
\begin{tcolorbox}[
    colback=white,
    colframe=black,
    boxrule=0.8pt,
    arc=0pt,
    left=4pt, right=4pt, top=4pt, bottom=4pt,
    label=alg:buffer
]
\small
\textbf{Algorithm 2: Buffer Management}\\
\vspace{0.1cm}
\textbf{Require:} Segment data $\mathcal{D}_k$, Buffer $\mathcal{B}$, Threshold $\tau$\\
\textbf{1:} Compute $\mathcal{L}_{inst}(x)$ for all instances $x \in \mathcal{D}_k$\\
\textbf{2:} \textbf{Hard Example Mining:}\\
\quad Identify $\mathcal{S}_{hard} \gets \{x \in \mathcal{D}_k \mid \mathcal{L}_{inst} > \tau \bar{\mathcal{L}}\}$\\
\textbf{3:} \textbf{Buffer Update:}\\
\quad $\mathcal{B}_{gv} \gets \text{Sample}(60\% \text{ from } \mathcal{S}_{hard}$\\
\quad \quad $\cup \ 40\% \text{ Random})$\\
\quad $\mathcal{B}_{gen} \gets \text{Sample}(300 \text{ Balanced from Kathbath})$\\
\textbf{4:} $\mathcal{B} \gets \mathcal{B}_{gv} \cup \mathcal{B}_{gen}$
\end{tcolorbox}
\end{figure}

\subsection{Computational Efficiency and Hardware Setup}
\label{sec:appendix_hardware}
A core goal of our work is to show that high-performance, on-device adaptation is feasible on consumer-grade laptops. We evaluated our pipeline on the configurations detailed in Table \ref{tab:hardware_specs}. These benchmarks demonstrate the technical feasibility of localized, self-improving ASR systems in environments where high-end compute clusters are unavailable. The reported runtimes are approximate, representative values and vary across software environments.

\begin{table}[t]
\centering
\small
\begin{tabularx}{\columnwidth}{@{}>{\raggedright\arraybackslash}p{2.1cm}>{\raggedright\arraybackslash}X>{\raggedright\arraybackslash}X@{}}
\toprule
 & \textbf{Config A (High-End)} & \textbf{Config B (Mid-Range)} \\
\midrule
CPU & Intel i7-13700H (14 cores) & Intel i5-12500H (12 cores) \\
GPU & NVIDIA RTX 4050 (35W) & NVIDIA RTX 3050 (70W) \\
Training Time (per segment) & 25--30 minutes & 50--60 minutes \\
\bottomrule
\end{tabularx}
\caption{Hardware configurations (consumer-grade laptops) and approximate training time benchmarks for on-device adaptation.}
\label{tab:hardware_specs}
\end{table}

\section{Dataset Characteristics}
\label{sec:appendix_data}

The Gram Vaani dataset serves as a proxy for rural clinical environments due to its telephonic acquisition (originally 8kHz upsampled to 16kHz) and focus on healthcare and agricultural discussions. Table \ref{tab:dataset_stats} summarizes the key characteristics. Forgetting is evaluated on the complete Kathbath Hindi validation set (3,151 samples) and the FLEURS Hindi (\texttt{hi\_in}) test split.

\begin{table}[t]
    \centering
    \small
    \begin{tabularx}{\columnwidth}{@{}l>{\raggedright\arraybackslash}X@{}}
        \toprule
        \textbf{Metric} & \textbf{Value} \\
        \midrule
        Total Duration & 103.2 Hours \\
        Number of Segments ($k$) & 19 (V1.1, V4.5); 24 (V2.1, V3.1, V5.1) \\
        New Samples per Segment & 2,000 (V1.1, V4.5); 1,600 (V2.1, V3.1, V5.1) \\
        Avg. Duration per Sample & 9.4 s \\
        Acoustic Condition & Noisy \\
        Primary Dialect & Rural Hindi \\
        \bottomrule
    \end{tabularx}
    \caption{Characteristics of the partitioned Gram Vaani dataset used for continual adaptation.}
    \label{tab:dataset_stats}
\end{table}

\section{Detailed Results by Paradigm}
\label{sec:appendix_results}

\begin{figure*}[p]
\centering
\includegraphics[width=0.9\textwidth]{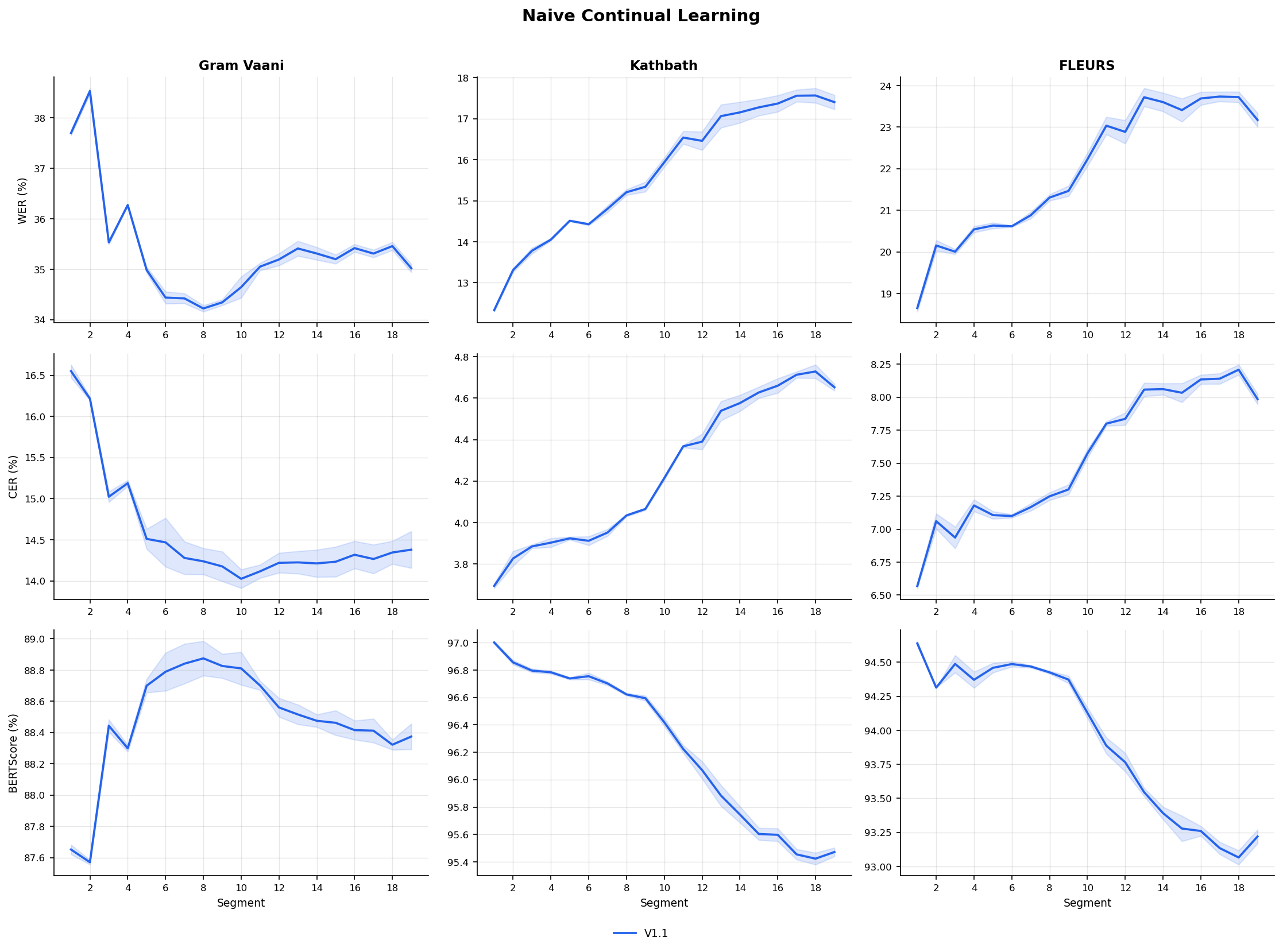}
\caption{Metric progression (WER, CER, BERTScore) for the Naive strategy (V1.1) across all three evaluation datasets. Shaded bands denote $\pm 1$ standard deviation across seeds.}
\label{fig:appendix_naive}
\end{figure*}

\begin{figure*}[p]
\centering
\includegraphics[width=0.9\textwidth]{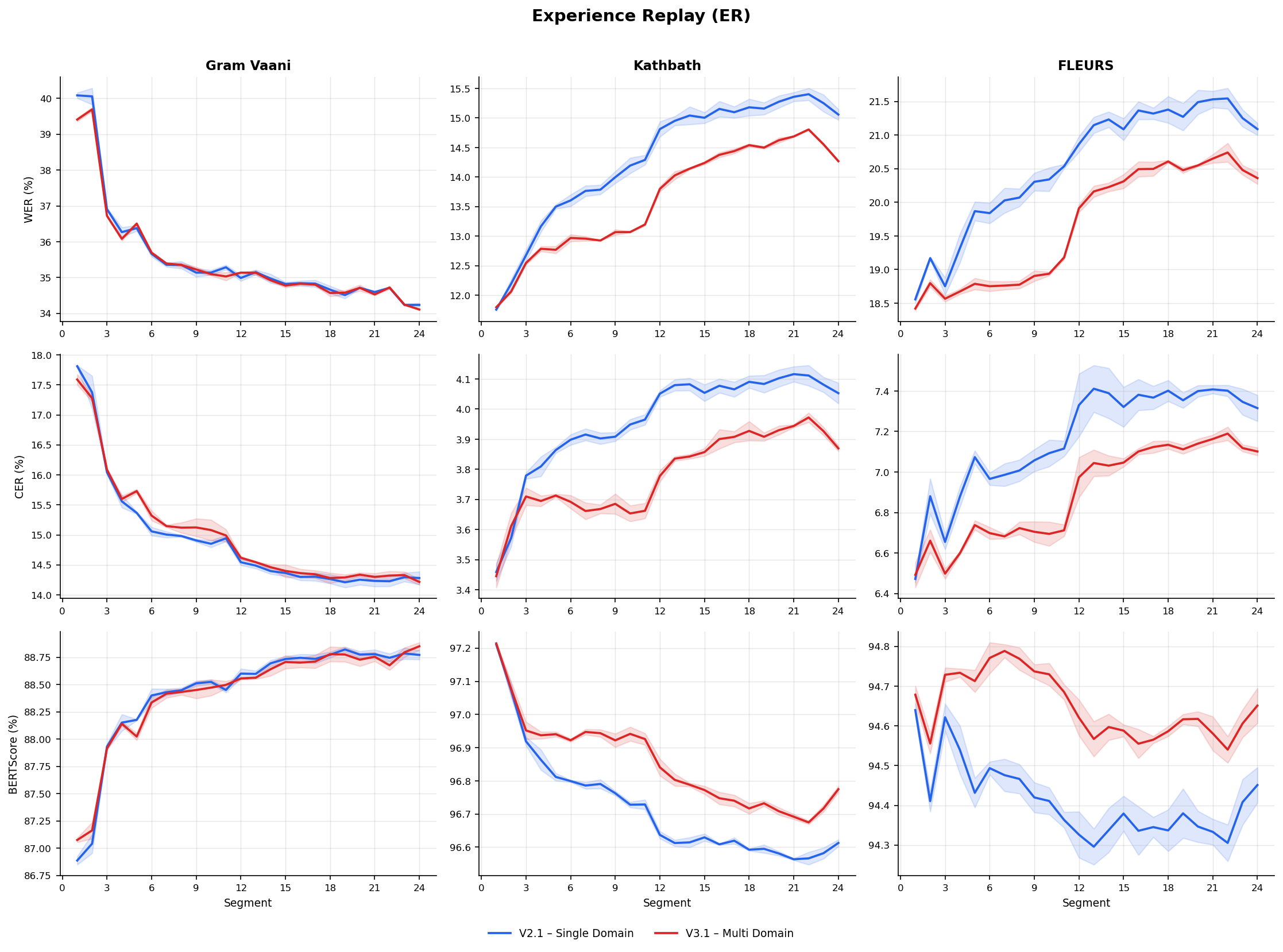}
\caption{Metric progression for Experience Replay strategies (V2.1 and V3.1) across all evaluation datasets. Shaded bands denote $\pm 1$ standard deviation across seeds.}
\label{fig:appendix_er}
\end{figure*}

\begin{figure*}[p]
\centering
\includegraphics[width=0.9\textwidth]{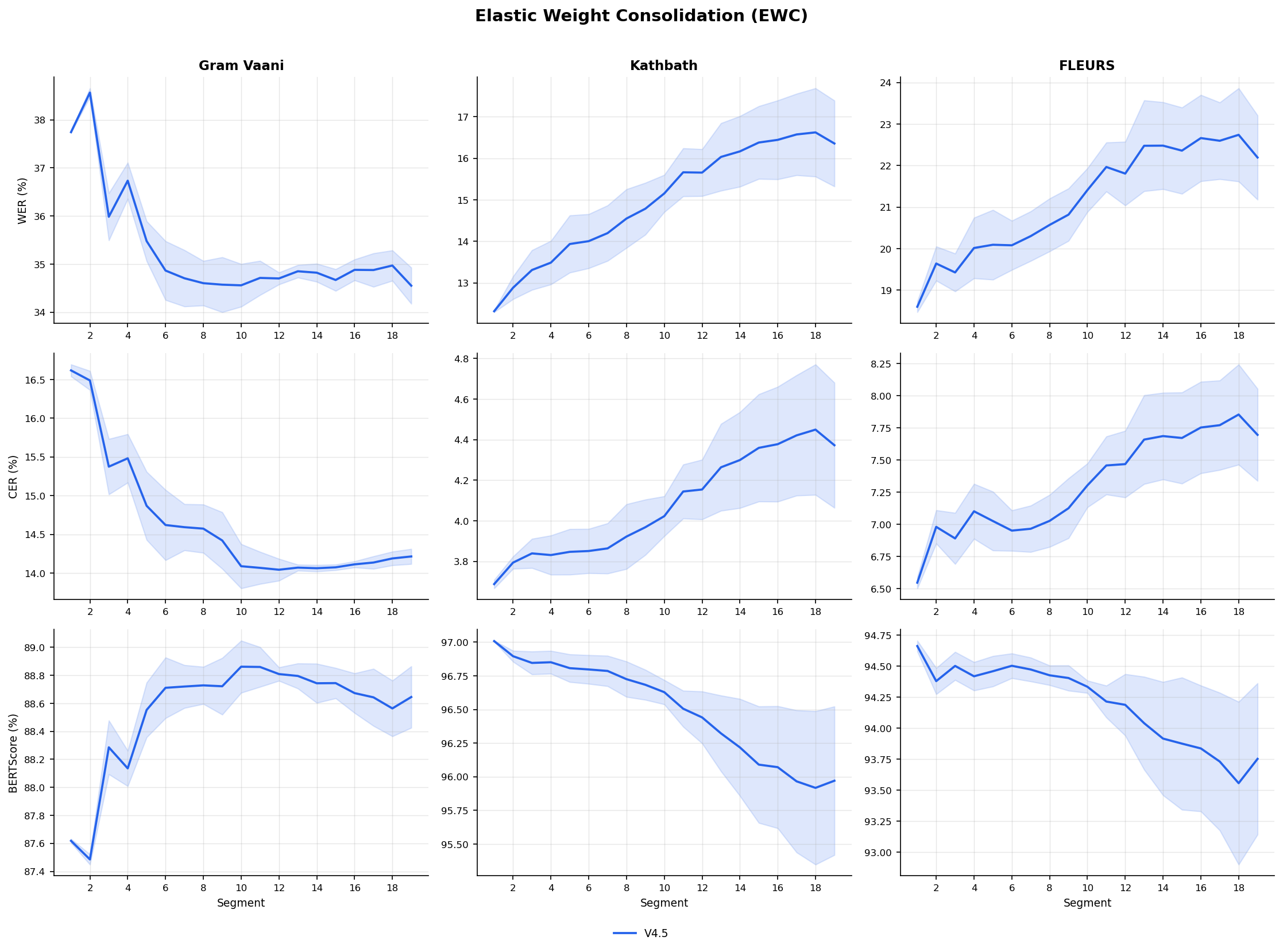}
\caption{Metric progression for the EWC strategy (V4.5). The larger standard deviation bands indicate higher sensitivity to the random seed than the replay-based strategies.}
\label{fig:appendix_ewc}
\end{figure*}

\begin{figure*}[p]
\centering
\includegraphics[width=0.9\textwidth]{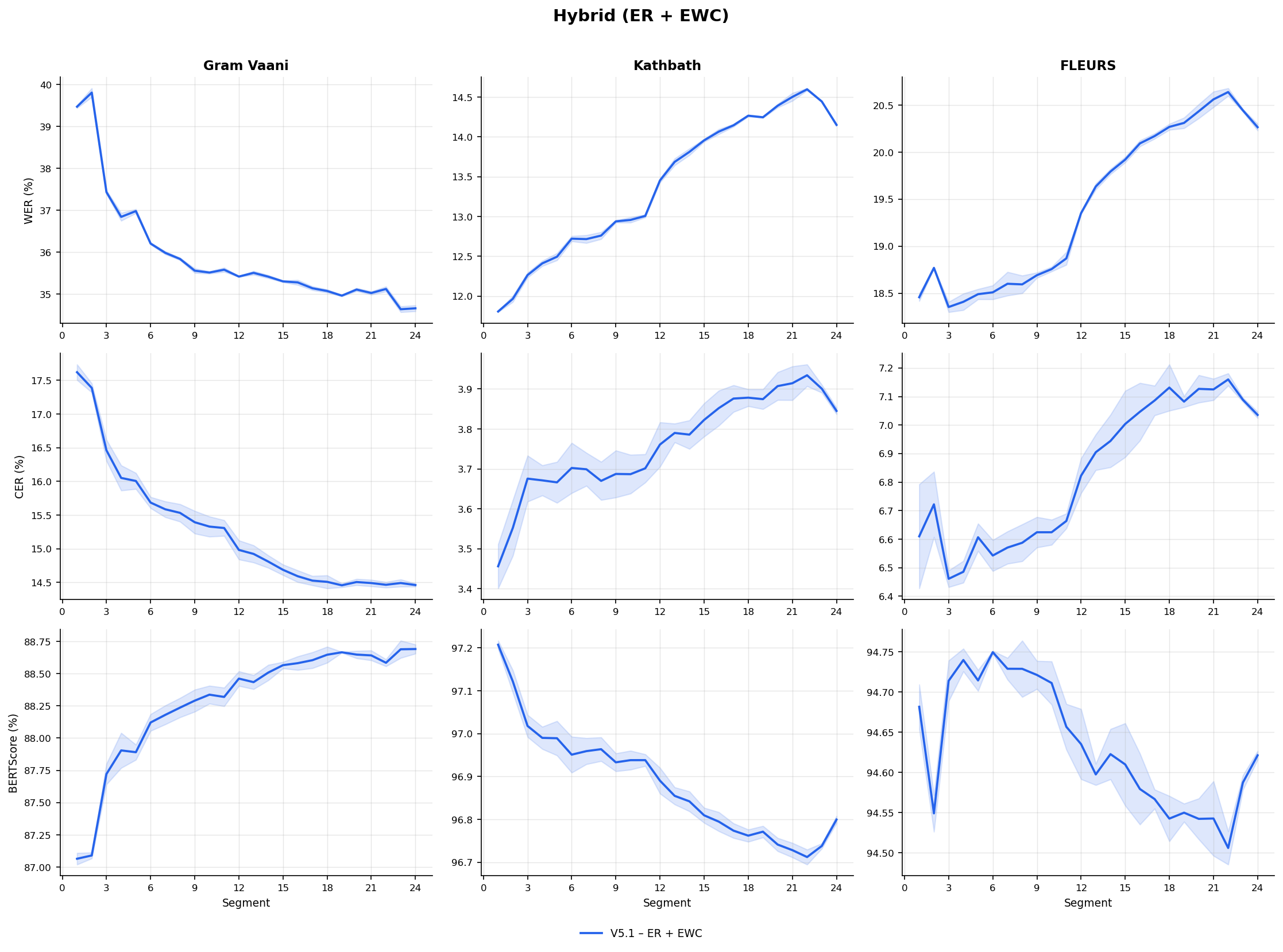}
\caption{Metric progression for the Hybrid strategy (V5.1) across all evaluation datasets. Shaded bands denote $\pm 1$ standard deviation across seeds.}
\label{fig:appendix_hybrid}
\end{figure*}

This section presents metric progressions for each continual strategy, grouped by paradigm: Naive (V1.1; Figure~\ref{fig:appendix_naive}), Experience Replay (V2.1 and V3.1; Figure~\ref{fig:appendix_er}), EWC (V4.5; Figure~\ref{fig:appendix_ewc}), and Hybrid (V5.1; Figure~\ref{fig:appendix_hybrid}). All metrics (WER, CER, BERTScore) are reported as percentages (\%) with mean $\pm$ standard deviation across 3 independent seeds, each run for the full continual adaptation protocol; final-checkpoint results for all strategies are reported in Table~\ref{tab:main_results}. Figures show the full progression over training segments; the shaded band denotes $\pm 1$ standard deviation. The EWC strategy (V4.5) shows visibly wider bands, indicating higher sensitivity to the random seed than the replay-based strategies.

\end{document}